# Efficient Smoothed Concomitant Lasso Estimation for High Dimensional Regression

Eugene Ndiaye[1], Olivier Fercoq[1], Alexandre Gramfort[1], Vincent Leclère[2], and Joseph Salmon[1]

[1]LTCI, CNRS, Télécom ParisTech, Université Paris-Saclay, 75013, Paris, France
[2]Université Paris-Est, Cermics (ENPC), F-77455 Marne-la-Vallée

June 8, 2016

## Abstract

In high dimensional settings, sparse structures are crucial for efficiency, both in term of memory, computation and performance. It is customary to consider $\ell_1$ penalty to enforce sparsity in such scenarios. Sparsity enforcing methods, the Lasso being a canonical example, are popular candidates to address high dimension. For efficiency, they rely on tuning a parameter trading data fitting versus sparsity. For the Lasso theory to hold this tuning parameter should be proportional to the noise level, yet the latter is often unknown in practice. A possible remedy is to jointly optimize over the regression parameter as well as over the noise level. This has been considered under several names in the literature: Scaled-Lasso, Square-root Lasso, Concomitant Lasso estimation for instance, and could be of interest for confidence sets or uncertainty quantification. In this work, after illustrating numerical difficulties for the Smoothed Concomitant Lasso formulation, we propose a modification we coined Smoothed Concomitant Lasso, aimed at increasing numerical stability. We propose an efficient and accurate solver leading to a computational cost no more expansive than the one for the Lasso. We leverage on standard ingredients behind the success of fast Lasso solvers: a coordinate descent algorithm, combined with safe screening rules to achieve speed efficiency, by eliminating early irrelevant features.

## 1 Related work

In the context of high dimensional regression where the number of features is greater than the number of observations, standard least squares need some regularization to both avoid over-fitting and ease the interpretation of discriminant features. Among the least squares with sparsity inducing regularization, the Lasso [27], using the $\ell_1$ norm as a regularizer, is the most standard one. It hinges on a regularization parameter governing the trade-off between data fitting and sparsity of the estimator, and requires careful tuning. Though this estimator is well understood theoretically, the choice of the tuning parameter remains an open and critical question in practice as well as in theory. For the Lasso, statistical guarantees [6] rely on choosing the tuning parameter proportional to the noise level, a quantity that is usually unknown to practitioners. Besides, the noise level is of practical interest since it is required in the computation of model selection criterions such as AIC, BIC, SURE or in the construction of confidence sets.

A convenient way to estimate both the regression coefficient and the noise level is to perform a joint estimation, for instance by performing the penalized maximum likelihood of the joint distribution. Unfortunately, a direct approach leads to a non-convex formulation (though, a change of variable can make it a jointly convex formulation [23]).

Another road for this joint estimation was inspired by the robust theory developed by Huber [15], particularly in the context of location-scale estimation. Indeed, Owen [19] extended it to handle sparsity inducing penalty, leading to a jointly convex optimization formulation. Since then, his estimator has appeared under various name, and we coined it the Concomitant Lasso. Indeed, as far as we know Owen was the first to propose such a formulation.

Later, the same formulation was mentioned in [1], in a response to [23], and was thoroughly analyzed in [25], under the name Scaled-Lasso. Similar results were independently obtained in [5] for the same estimator, though with a different formulation. While investigating pivotal quantities, Belloni *et al.* proposed to solve the following convex program: modify the standard Lasso by removing the square in the data fitting term. Thus, they termed their estimator the Square-root Lasso (see also [7]). A second approach leading to this very formulation, was proposed by [28] to account for noise in the design matrix, in an adversarial scenario. Interestingly their robust construction led exactly to the Square-root Lasso formulation.

Under standard design assumption (see [6]), it is proved that the Scaled/Square-root Lasso reaches optimal rates for sparse regression, with the additional benefit that the regularization parameter is independent of the noise level [5, 25]. Moreover, a practical study [22] has shown that the Concomitant Lasso estimator, or its debiased version (see for instance [4, 16] for a discussion on least-squares refitting), is particularly well suited for estimating the noise level in high dimension.

Among the solutions to compute the Concomitant Lasso, two roads have been pursued. On the one hand, considering the Scaled-Lasso formulation, Sun and Zhang [24, 25] have proposed an iterative procedure that alternates Lasso steps and noise estimation steps, the later leading to rescaling the tuning parameter iteratively. On the other hand, considering the Square-root Lasso formulation, Belloni *et al.* [5] have leaned on second order cone programming solvers, *e.g.*, TFCOS [3].

Despite the appealing properties listed above, among which the superiority of the theoretical results is the most striking, no consensus for an efficient solver has yet emerged for the Concomitant Lasso. Our contribution aims at providing a more numerically stable formulation, called the Smoothed Concomitant Lasso. This variant allows to obtain a fast solver: we first adapt a coordinate descent algorithm to the smooth version (in the sense given in [18]) of the original problem. Then, we apply safe rules strategies, introduced in [10] to our estimator. Such rules allow to discard features whose coefficients are certified to be zero, either prior any computation or as the algorithm proceeds. Combined with a coordinate descent, this leads to important acceleration in practice, as illustrated for the Lasso case [13]. We show similar accelerations for the Smoothed Concomitant Lasso, both on real and simulated data. Moreover, leveraging such screening rules and the active sets they provide, we have introduced a novel warm start strategy offering additional speed-ups. Overall, our method presents the same computational cost as for the Lasso, but enjoys the nice features mentioned earlier in terms of statistical properties.

## 2 Concomitant estimator

In the following we present our estimator and some important properties. All the proof are deferred to the Appendix due to space constraint.

**Notation** For any integer $d \in \mathbb{N}$, we denote by $[d]$ the set $\{1,\ldots,d\}$. Our observation vector is $y \in \mathbb{R}^n$ (note that for simplicity, we assume that the observed signal $y$ is nonzero $y \neq 0$) and the design matrix $X = [X_1,\ldots,X_p] \in \mathbb{R}^{n \times p}$ has $p$ explanatory variables or features, stored columnwise. The standard Euclidean norm is written $\|\cdot\|$, the $\ell_1$ norm $\|\cdot\|_1$, the $\ell_\infty$ norm $\|\cdot\|_\infty$, and the matrix transposition of a matrix $Q$ is denoted by $Q^\top$. We note $\mathcal{B}_\infty$ the unit ball with the $\ell_\infty$ norm. For real numbers $a$ and $b$, we write $a \vee b$ for the maximum of $a$ and $b$.

We denote $\mathcal{S}_\tau$ the soft-thresholding operator at level $\tau > 0$, i.e., for $x \in \mathbb{R}$, $\mathcal{S}_\tau(x) = \text{sign}(x)(|x| - \tau)_+$. For a closed convex set $\mathcal{C}$, we write $\Pi_\mathcal{C}$ the projection over the set $\mathcal{C}$. The sub-gradient of a convex function $f : \mathbb{R}^d \to \mathbb{R}$ at $x$ is defined as $\partial f(x) = \{z \in \mathbb{R}^d : \forall y \in \mathbb{R}^d, f(x) - f(y) \geq z^\top(x-y)\}$. We denote by $f^*$ the Fenchel-conjugate of $f$, $f^*(z) = \sup_{w \in \mathbb{R}^d} \langle w, z \rangle - f(w)$ and by $\iota_C$ the indicator function of a set $C$ defined as $\iota_C(x) = 0$ if $x \in C$ and $\iota_C(x) = \infty$ if $x \notin C$.

We recall that the sub-differential $\partial \|\cdot\|_1$ of the $\ell_1$ norm is the set-valued function $\text{sign}(\cdot)$, defined element-wise for all $j \in [d]$ by $\text{sign}(x_j) = 1$ if $x_j > 0$, by $\text{sign}(x_j) = 1$ if $x_j < 0$ and by $\text{sign}(x_j) = [-1, 1]$ if $x_j = 0$.

For a set $S \subset [p]$, we denote by $P_{X,S} = X_S \left(X_S^\top X_S\right)^+ X_S^\top$ the projection operator onto $\text{Span}\{X_j : j \in S\}$, where $A^+$ represents the Moore-Penrose pseudo-inverse.

We note $\text{tr}(X)$ the trace of matrix $X$ and $\widehat{\Sigma} = X^\top X / n$ the normalized Gram matrix of $X$.

### 2.1 Concomitant Lasso

Let us first introduce the Concomitant Lasso estimator, following the formulation proposed in [19, 25], and present some properties obtained due to convexity and duality.

**Definition 1.** For $\lambda > 0$, the Concomitant Lasso estimator $\hat{\beta}^{(\lambda)}$ is defined as a solution of the primal optimization problem

$$(\hat{\beta}^{(\lambda)}, \hat{\sigma}^{(\lambda)}) \in \underset{\beta \in \mathbb{R}^p, \sigma > 0}{\arg\min} \underbrace{\frac{1}{2n\sigma} \|y - X\beta\|^2 + \frac{\sigma}{2} + \lambda \|\beta\|_1}_{P_\lambda(\beta,\sigma)}, \tag{1}$$

**Theorem 1.** Denoting $\Delta_{X,\lambda} = \{\theta \in \mathbb{R}^n : \|X^\top \theta\|_\infty \leq 1, \lambda\sqrt{n}\|\theta\| \leq 1\}$, the dual formulation of the Concomitant Lasso reads

$$\hat{\theta}^{(\lambda)} \in \underset{\theta \in \Delta_{X,\lambda}}{\arg\max} \underbrace{\langle y, \lambda\theta \rangle}_{D_\lambda(\theta)}. \tag{2}$$

For an optimal primal vector $\hat{\beta}^{(\lambda)}$, $\hat{\sigma}^{(\lambda)} = \|y - X\hat{\beta}^{(\lambda)}\|/\sqrt{n}$. Moreover, the Fermat's rule reads

$$y = n\lambda\hat{\sigma}^{(\lambda)}\hat{\theta}^{(\lambda)} + X\hat{\beta}^{(\lambda)} \text{ \textbf{\textit{(link-equation)}}}, \tag{3}$$

$$X^\top(y - X\hat{\beta}^{(\lambda)}) \in n\lambda\hat{\sigma}^{(\lambda)}\partial\|\cdot\|_1(\hat{\beta}^{(\lambda)}) \text{ \textbf{\textit{(sub-differential inclusion)}}}. \tag{4}$$

**Remark 1.** As defined in (1), the Concomitant Lasso estimator is ill-defined. Indeed, the set over which we optimize is not closed and the optimization problem may have no solution. We circumvent this difficulty by considering instead the Fenchel biconjugate of the objective function (for more details, see Appendix C). The actual objective function accepts $\sigma \geq 0$ as soon as $y = X\beta$. In the rest of the paper, we will write (1) instead of the minimization of the biconjugate as a slight abuse of notation.

**Remark 2.** The Square-root Lasso formulation [5] is expressed as

$$\hat{\beta}^{(\lambda)}_{\sqrt{\text{Lasso}}} \in \arg\min_{\beta \in \mathbb{R}^p} \frac{1}{\sqrt{n}} \|y - X\beta\| + \lambda \|\beta\|_1 . \tag{5}$$

It can be verified that $\left(\hat{\beta}^{(\lambda)}_{\sqrt{\text{Lasso}}}, \hat{\sigma}^{(\lambda)}_{\sqrt{\text{Lasso}}}\right)$ where, $\hat{\sigma}^{(\lambda)}_{\sqrt{\text{Lasso}}} = \|y - X\hat{\beta}^{(\lambda)}_{\sqrt{\text{Lasso}}}\|/\sqrt{n}$, is a solution of the Concomitant Lasso (1) for all $\lambda > 0$.

Since it is difficult to get the right regularization parameter in advance, a principled way to tune Lasso-type programs is to perform a cross-validation procedure over a pre-set finite grid of parameters. This leads to a data-driven choice of regularizer requiring the computation of many estimators, one for each $\lambda$ value. Usually, a geometrical grid $\lambda_t = \lambda^{\text{L}}_{\max} 10^{-\delta(t-1)/(T-1)}, t \in [T]$ is used, for instance it is the default grid in `scikit-learn` [11] and `glmnet` [14], with $\delta = 3$.

For the Concomitant Lasso, we now show that this method presents some numerical drawbacks. Let us first investigate the solution for extreme values of $\lambda$.

## 2.2 Critical parameters for the Concomitant Lasso

As for the Lasso, the null vector is optimal for the Concomitant Lasso problem as soon as the regularization parameter becomes too large, as detailed in the next proposition.

**Proposition 1.** *For all $\lambda \geq \lambda_{\max} := \|X^\top y\|_\infty / (\|y\|\sqrt{n})$, we have $\hat{\beta}^{(\lambda)} = 0$.*

*Proof.* The Fermat's rule states:

$$(0, \hat{\sigma}^{(\lambda)}) \in \arg\min_{\beta \in \mathbb{R}^p, \sigma > 0} P_\lambda(\beta, \sigma) \Longleftrightarrow 0 \in \frac{\{-X^\top y\}}{n\hat{\sigma}^{(\lambda)}} + \lambda \mathcal{B}_\infty \Longleftrightarrow \frac{1}{n\hat{\sigma}^{(\lambda)}} \|X^\top y\|_\infty \leq \lambda.$$

Thus, the critical parameter is given by $\lambda_{\max} = \|X^\top y\|_\infty / (n\hat{\sigma}^{(\lambda)})$, so noticing that when $\hat{\beta}^{(\lambda)} = 0$ one has $\hat{\sigma}^{(\lambda)} = \|y\|/\sqrt{n} > 0$ (remind that we assumed $y \neq 0$) the results follows. □

However, for the Concomitant Lasso, there is another extreme. Indeed, there exists a critical parameter $\lambda_{\min}$ such that the Concomitant Lasso is equivalent to the Basis Pursuit for all $\lambda \leq \lambda_{\min}$ and gives an estimate $\hat{\sigma}^{(\lambda)} = 0$. We recall that the Basis Pursuit and its dual are given by

$$\hat{\beta}^{\text{BP}} \in \arg\min_{\beta \in \mathbb{R}^p : y = X\beta} \|\beta\|_1 , \quad \hat{\theta}^{\text{BP}} \in \arg\max_{\theta \in \mathbb{R}^n : \|X^\top \theta\|_\infty \leq 1} \langle y, \theta \rangle . \tag{6}$$

**Proposition 2.** *For any $\hat{\theta}^{\text{BP}} \in \arg\max_{\theta \in \mathbb{R}^n : \|X^\top \theta\|_\infty \leq 1} \langle y, \theta \rangle$ and any $\lambda \leq \lambda_{\min} := 1/(\|\hat{\theta}^{\text{BP}}\|\sqrt{n})$, $(\hat{\beta}^{\text{BP}}, 0)$ is optimal for $P_\lambda$ and $\hat{\theta}^{\text{BP}}$ is optimal for $D_\lambda$.*

*Proof.* By strong duality in the Basis Pursuit problem $\|\hat\beta^{\text{BP}}\|_1 = \langle y, \hat\theta^{\text{BP}}\rangle$. Now, $(\hat\beta^{\text{BP}}, 0)$ is admissible for $P_\lambda$ (see Remark 1) and $\hat\theta^{\text{BP}}$ is admissible for $D_\lambda$ as soon as $\lambda \leq \lambda_{\min} = 1/(\|\hat\theta^{\text{BP}}\|\sqrt{n})$. One can check for $\lambda \leq \lambda_{\min}$ that $P_\lambda(\hat\beta^{\text{BP}}, 0) = \lambda\|\hat\beta^{\text{BP}}\|_1 = \lambda\langle y, \hat\theta^{\text{BP}}\rangle = D_\lambda(\hat\theta^{\text{BP}})$. We conclude that $(\hat\beta^{\text{BP}}, 0)$ is optimal for the primal and $\hat\theta^{\text{BP}}$ is optimal for the dual. $\square$

We can guarantee the existence of minimizers to the Concomitant Lasso (see Appendix C), even if $\hat\sigma^{(\lambda)} = 0$, but the problem becomes more and more ill-conditioned for smaller and smaller $\hat\sigma^{(\lambda)}$. The previous proposition shows that for too small $\lambda$'s, a Basis Pursuit solution will always be found, though numerically this might be challenging to get.

Indeed, when $\lambda$ approaches $\lambda_{\min}$, a coordinate descent algorithm (similar to the one described in Algorithm 1) encounters trouble to perform dual gap computations. This is because we estimate the dual variable by a ratio having both denominator and numerator of the order of $\sigma$, which is problematic when $\sigma \to 0$ (*cf.* Eq. (9)).

A solution could be to pre-compute $\lambda_{\min}$ to prevent the user from requesting computation involving $\lambda$'s too close from the critical value. Nevertheless, solving the Basis Pursuit problem first, to obtain $\lambda_{\min}$, is not realistic. This step is the most difficult one to solve on the path of $\lambda$'s, and in such a case one would not benefit from warm start computations.

To avoid these issues, we propose a slight modification of the objective function by adding a constraint on $\sigma$. We refer to this method as the Smoothed Concomitant Lasso following the terminology introduced by Nesterov [18].

## 2.3 Smoothed Concomitant Lasso

We now introduce our Smoothed Concomitant Lasso, by adding a noise level limit $\sigma_0$, aimed at avoiding numerical instabilities for too small $\lambda$ values.

**Definition 2.** *For $\lambda > 0$ and $\sigma_0 > 0$, the Smoothed Concomitant Lasso estimator $\hat\beta^{(\lambda,\sigma_0)}$ and its associated noise level estimate $\hat\sigma^{(\lambda,\sigma_0)}$ are defined as solutions of the primal optimization problem*

$$(\hat\beta^{(\lambda,\sigma_0)}, \hat\sigma^{(\lambda,\sigma_0)}) \in \underset{\beta\in\mathbb{R}^p, \sigma\in\mathbb{R}}{\arg\min} \underbrace{\frac{1}{2n\sigma}\|y - X\beta\|^2 + \frac{\sigma}{2} + \lambda\|\beta\|_1 + \iota_{[\sigma_0, +\infty[}(\sigma)}_{P_{\lambda,\sigma_0}(\beta,\sigma)}. \tag{7}$$

**Theorem 2.** *With $\Delta_{X,\lambda} = \{\theta \in \mathbb{R}^n : \|X^\top\theta\|_\infty \leq 1, \|\theta\| \leq 1/(\lambda\sqrt{n})\}$, the dual formulation of the Smoothed Concomitant Lasso reads*

$$\hat\theta^{(\lambda,\sigma_0)} = \underset{\theta\in\Delta_{X,\lambda}}{\arg\max} \underbrace{\langle y, \lambda\theta\rangle + \sigma_0\left(\frac{1}{2} - \frac{\lambda^2 n}{2}\|\theta\|^2\right)}_{D_{\lambda,\sigma_0}(\theta)}. \tag{8}$$

*For an optimal primal vector $\hat\beta^{(\lambda,\sigma_0)}$, we must have $\hat\sigma^{(\lambda,\sigma_0)} = \sigma_0 \vee (\|y - X\hat\beta^{(\lambda,\sigma_0)}\|/\sqrt{n})$. We also have the link-equation between primal and dual solutions: $y = n\lambda\hat\sigma^{(\lambda,\sigma_0)}\hat\theta^{(\lambda,\sigma_0)} + X\hat\beta^{(\lambda,\sigma_0)}$ and the sub-differential inclusion $X^\top(y - X\hat\beta^{(\lambda,\sigma_0)}) \in n\lambda\hat\sigma^{(\lambda,\sigma_0)}\partial\|\cdot\|_1(\hat\beta^{(\lambda,\sigma_0)})$.*

**Remark 3.** *Problem (8) also reads $\hat\theta^{(\lambda,\sigma_0)} = \arg\max_{\theta\in\Delta_{X,\lambda}} \|y/(\sigma_0 n)\|^2/2 - \|\lambda\theta - y/(\sigma_0 n)\|^2/2 = \Pi_{\Delta_{X,\lambda}}(y/(\lambda\sigma_0 n))$. Since $\Delta_{X,\lambda}$ is convex and closed, $\hat\theta^{(\lambda,\sigma_0)}$ is unique.*

The choice of $\sigma_0$ can be motivated as follows:

- Suppose we have prior information on the minimal noise level expected in the data. Then we can set $\sigma_0$ as this bound. Indeed, if $\hat{\sigma}^{(\lambda,\sigma_0)} > \sigma_0$, then the constraint $\sigma \geq \sigma_0$ is not active and the optimal solution to Problem (7) is equal to the optimal solution to Problem (1). The Smoothed Concomitant Lasso estimator will only be different from the Concomitant Lasso estimator when the prediction given by the Concomitant Lasso violates the a priori information.

- Without prior information we can consider a given accuracy $\epsilon$, and set $\sigma_0 = \epsilon$. Then, the theory of smoothing [18] tells us that any $\epsilon/2$-solution to Problem (7) is an $\epsilon$-solution to Problem (1). Thus we obtain the same solutions, but as an additional benefit we have a control on the conditioning of the problem.

- One can also use a proportion of the initial estimation of the noise variance *i.e.*, $\sigma_0 = \|y\|/\sqrt{n} \times 10^{-\alpha}$. This was our choice in practice, and we have sets $\alpha = 2$.

A similar reasoning to Proposition 1 gives the following critical parameter.

**Proposition 3.** *For all* $\lambda \geq \lambda_{\max} := \|X^\top y\|_\infty/(n\,(\sigma_0 \vee (\|y\|/\sqrt{n})))$, *we have* $\hat{\beta}^{(\lambda,\sigma_0)} = 0$.

## 2.4 Coordinate descent, duality gap and link with the Lasso

We present the algorithm we consider to compute the Smoothed Concomitant Lasso: coordinate descent, an efficient way to solve Lasso-type problem (even for multiple values of parameters) [14]. Though it seems natural to consider such a method for the Concomitant Lasso, previous attempts mainly focused on iteratively alternating Lasso steps along with noise level estimation [25][1], or conic programming [3]. Here we provide a simple coordinate descent approach, *cf.* Algorithm 1. Our primal objective $P_{\lambda,\sigma_0}$ can be written as the sum of a convex differentiable function $f(\beta,\sigma) = \|y - X\beta\|^2/(2n\sigma) + \sigma/2$ and of a separable function $g(\beta,\sigma) = \lambda\|\beta\|_1 + \iota_{[\sigma_0,+\infty[}(\sigma)$. Moreover, for $\sigma \geq \sigma_0 > 0$, the gradient of $f$ is Lipschitz continuous. Hence, we know that the coordinate descent method converges to a minimizer of our problem [29]. We choose to update the variable $\sigma$ every other iteration because this can be done at a negligible cost.

Our stopping criterion is based on the duality gap defined by $G_{\lambda,\sigma_0}(\beta,\sigma,\theta) = P_{\lambda,\sigma_0}(\beta,\sigma) - D_{\lambda,\sigma_0}(\theta)$. This requires the computation of a dual feasible point, that, provided a primal vector $\beta$, can be obtained as follows

$$\theta = (y - X\beta)/\left(\lambda n\sigma_0 \vee \left\|X^\top(y-X\beta)\right\|_\infty \vee \lambda\sqrt{n}\,\|y - X\beta\|\right). \tag{9}$$

**Proposition 4.** *Let* $(\beta_k)_{k\in\mathbb{N}}$ *be a sequence that converges to* $\hat{\beta}^{(\lambda,\sigma_0)}$. *Then* $(\theta_k)_{k\in\mathbb{N}}$ *built thanks to* (9) *converges to* $\hat{\theta}^{(\lambda,\sigma_0)}$. *Hence the sequence of dual gap* $(G_{\lambda,\sigma_0}(\beta_k,\sigma_k,\theta_k))_{k\in\mathbb{N}}$ *converges to zero.*

From the optimality condition in (3) and (4), one can remark that if $\hat{\beta}^{(\lambda,\sigma_0)}$ is a solution of the Smoothed Concomitant Lasso, then it is also a solution of the Lasso with regularization parameter $\lambda\hat{\sigma}^{(\lambda,\sigma_0)}$. The following proposition estimates the quality (in term of duality gap) of a primal-dual vector in the Lasso path compared to Concomitant Lasso path. We recall the Lasso problem and its dual

$$\hat{\beta}_{\mathrm{L}}^\lambda \in \underset{\beta\in\mathbb{R}^p}{\arg\min}\ \underbrace{\frac{1}{2n}\|y - X\beta\|^2 + \lambda\|\beta\|_1}_{P_\lambda^{\mathrm{L}}(\beta)},\quad \hat{\theta}_{\mathrm{L}}^\lambda = \underset{\theta\in\mathbb{R}^n:\|X^\top\theta\|_\infty \leq 1}{\arg\max}\ \underbrace{\frac{1}{2n}\|y\|^2 - \frac{1}{2n}\|y - \lambda n\theta\|^2}_{D_\lambda^{\mathrm{L}}(\theta)}.$$

---

[1] a description of their algorithm is given in Appendix A for completeness

**Algorithm 1:** Coordinate descent for the Smoothed Concomitant Lasso

**Input** : $X, y, \epsilon, K, f^{\text{ce}}, (\lambda_t)_{t \in [T-1]}, \sigma_0$
$\lambda_0 = \lambda_{\max} = \|X^\top y\|_\infty / (\|y\| \sqrt{n}), \quad \beta^{\lambda_0} = 0, \quad \sigma^{\lambda_0} = \|y\| / \sqrt{n}$
**for** $t \in [T-1]$ **do**
  $\beta, \sigma \leftarrow \beta^{\lambda_{t-1}}, \sigma^{\lambda_{t-1}}$ (previous $\epsilon$-solution)   // Get previous $\epsilon$-solution
  **for** $k \in [K]$ **do**
    **if** $k \mod f^{\text{ce}} = 1$ **then**
      Compute $\theta$ thanks to (9)
      **if** $G_{\lambda, \sigma_0}(\beta, \sigma, \theta) = P_{\lambda_t, \sigma_0}(\beta, \sigma) - D_{\lambda_t}(\theta) \leq \epsilon$ **then**   // Stopping criterion
        $\beta^{\lambda_t}, \sigma^{\lambda_t} \leftarrow \beta, \sigma$
        **break**
    **for** $j \in [p]$ **do**   // Loop over coordinate
      $\beta_j \leftarrow \mathcal{S}_{n\sigma\lambda_t / \|X_j\|^2} \left( \beta_j - X_j^\top (X\beta - y) / \|X_j\|^2 \right)$   // Soft-thresholding step
      $\sigma \leftarrow \sigma_0 \vee (\|y - X\beta\| / \sqrt{n}))$   // Noise estimation step
**Output**: $(\beta^{\lambda_t})_{t \in [T-1]}, (\sigma^{\lambda_t})_{t \in [T-1]}$

Hence, defining the dual gap of the Lasso $G_\lambda^{\text{L}}(\beta, \theta) = P_\lambda^{\text{L}}(\beta) - D_\lambda^{\text{L}}(\theta)$, one can easily show that

**Proposition 5.** $\forall \beta \in \mathbb{R}^p, \theta \in \Delta_{X,\lambda}, \sigma \geq \sigma_0, \quad G_{\sigma\lambda}^{\text{L}}(\beta, \theta) \leq \sigma G_{\lambda, \sigma_0}(\beta, \sigma, \theta).$

Hence, as $\forall \lambda, \hat{\sigma}^{(\lambda)} \leq \|y\| / \sqrt{n}$, if the duality gap for the Smoothed Concomitant Lasso is small, so is the duality gap for the Lasso with the corresponding regularization parameter.

## 3 Safe screening rules

In order to achieve a greater computational efficiency, we propose different Safe screening rules and we compare their performance. Following the seminal work of [10], one can discard inactive features, thanks to the sub-differential inclusion and to a safe region $\mathcal{R}$ such that $\hat{\theta}^{(\lambda, \sigma_0)} \in \mathcal{R}$:

$$\max_{\theta \in \mathcal{R}} |X_j^\top \theta| < 1 \implies |X_j^\top \hat{\theta}^{(\lambda, \sigma_0)}| < 1 \implies \hat{\beta}_j^{(\lambda, \sigma_0)} = 0. \tag{10}$$

Since the dual objective of the Smoothed Concomitant Lasso is $\lambda^2 \sigma_0 n$-strongly concave, we have from [17], a dynamic and converging SAFE sphere region $\mathcal{R}$.

**Proposition 6** (Gap Safe rule). *For all* $(\beta, \sigma, \theta) \in \mathbb{R}^p \times \mathbb{R}_+ \times \Delta_{X, \lambda}$, *then for* $r = \sqrt{2 G_{\lambda, \sigma_0}(\beta, \sigma, \theta) / (\lambda^2 \sigma_0 n)}$, *we have* $\hat{\theta}^{(\lambda, \sigma_0)} \in \mathcal{B}(\theta, r)$. *Thus, we have the following safe sphere screening rule*

$$|X_j^\top \theta| + r \|X_j\| < 1 \implies \hat{\beta}_j^{(\lambda, \sigma_0)} = 0. \tag{11}$$

Another test, valid when $\sigma_0 = 0$, can be derived if we assume upper/lower bounds: to eliminate feature $j$, it is enough to check whether $\max_\theta \{|X_j^\top \theta| : \lambda \sqrt{n} \|\theta\| \leq 1, \quad \underline{\eta} \leq D_\lambda(\theta) \leq \bar{\eta}\} < 1$.

In our implementation, we use the primal and the dual objective as a natural bound on the problem : indeed, $\underline{\eta} = D_\lambda(\theta_k) \leq D_\lambda(\hat{\theta}^{(\lambda, \sigma_0)}) \leq P_{\lambda, \sigma_0}(\beta_k, \sigma_k) = \bar{\eta}$.

**Proposition 7** (Bound Safe rule). *Assume that, for a given* $\lambda > 0$, *we have an upper bound* $\bar{\eta} \in (0, +\infty]$, *and a lower bound* $\underline{\eta} \in (0, +\infty]$ *over the Smoothed Concomitant Lasso problem* (7).

Denote by $x_j = X_j/\|X_j\|$ and $y' = y/\|y\|$ two unit vectors, and by $\underline{\gamma} = (\underline{\eta} - \sigma_0/2)\sqrt{n}/\|y\|$ and $\overline{\gamma} = \bar{\eta}\sqrt{n}/\|y\|$. Then if one of the three following conditions is met

- $|x_j^\top y'| > \overline{\gamma}$ and $\overline{\gamma}|x_j^\top y'| + \sqrt{1-\overline{\gamma}^2}\sqrt{1-(x_j^\top y')^2} < \lambda\sqrt{n}/\|X_j\|$,

- $\underline{\gamma} \leq |x_j^\top y'| \leq \overline{\gamma}$ and $1 < \lambda\sqrt{n}/\|X_j\|$,

- $|x_j^\top y'| < \underline{\gamma}$ and $\underline{\gamma}|x_j^\top y'| + \sqrt{1-\underline{\gamma}^2}\sqrt{1-(x_j^\top y')^2} < \lambda\sqrt{n}/\|X_j\|$,

the $j$-th feature can be discarded i.e., $\hat{\beta}_j^{(\lambda,\sigma_0)} = 0$.

## 4 Numerical experiments

We compare the estimation performance and computation times of standard deviation estimators which are presently the state of the art in high dimensional settings. We refer to [22] for a recent comparison. In our simulations we use the common setup: $y = X\beta^\star + \sigma^\star\varepsilon$ where $\varepsilon \sim \mathcal{N}(0, \mathrm{Id}_n)$ and $X \in \mathbb{R}^{n \times p}$ follows a multivariate normal distribution with covariance $\Sigma = (\rho^{|i-j|})_{i,j\in[p]}$. We define $\beta^\star = \alpha\beta$ where the coordinates of $\beta$ are drawn from a standard Laplace distribution and we randomly set $s\%$ of them to zero. The scalar $\alpha$ is chosen in order to satisfy a prescribed signal to noise ratio denoted snr: $\alpha = \sqrt{\mathrm{snr} \times \sigma^2/\beta^\top\Sigma\beta}$. We note $S^\star = \{j \in [p], \beta_j^\star \neq 0\}$.

We briefly recall the procedures we have compared. Our reference is the oracle estimator (OR) $\hat{\sigma}_{OR} = \|y - P_{X,S^\star}y\|/(n - |S^\star|)^{1/2}$ (note that this is a "theoretical" estimator, since it requires the knowledge of the true support $S^\star$ to be computed). We denote $\hat{\beta}_{\mathcal{M}}$ an estimator obtained by a method $\mathcal{M}$. To obtain the cross-validation estimator (CV), we first pick the parameter $\lambda_{cv}$ by 5-fold cross-validation and define $\hat{\sigma}_{\mathcal{M}-CV} = \|y - X\hat{\beta}_{\mathcal{M}}^{\lambda_{cv}}\|/(n - |\hat{S}_{\mathcal{M}}^{\lambda_{cv}}|)^{1/2}$ where $\hat{\beta}_{\mathcal{M}}^{\lambda_{cv}}$ is obtained by using all the data, and $\hat{S}_{\mathcal{M}}^{\lambda_{cv}} = \{j \in [p], \hat{\beta}_{\mathcal{M}}^{\lambda_{cv}} \neq 0\}$. The least-square refitting estimator (LS) is $\hat{\sigma}_{\mathcal{M}-LS} = \|y - P_{X,\hat{S}_{\mathcal{M}}}y\|/(n - |\hat{S}_{\mathcal{M}}|)^{1/2}$. For the refitted cross-validation (RCV) method the dataset is split in two parts ($D_i = (y^{(i)}, X^{(i)})_{i\in[2]}$). Let $\hat{S}_i$ be the support selected after a cross-validation on the part $D_i$. Then define $\hat{\sigma}_{RCV} = ((\hat{\sigma}_1^2 + \hat{\sigma}_2^2)/2)^{1/2}$ where $\hat{\sigma}_1 = \|y^{(2)} - P_{X^{(2)},\hat{S}_1}y^{(2)}\|/(n/2 - |\hat{S}_1|)^{1/2}$. The value of $\hat{\sigma}_2$ is obtained by swapping 1 and 2 in the last formula. Finally [9] introduces the estimator $\hat{\sigma}_{D2} = ((1 + p\hat{m}_1^2/((n+1)\hat{m}_2))\|y\|^2/n - \hat{m}_1\|X^\top y\|^2/(n(n+1)\hat{m}_2))^{1/2}$ where $\hat{m}_1 = \mathrm{tr}(\hat{\Sigma})/p$ and $\hat{m}_2 = \mathrm{tr}(\hat{\Sigma}^2)/p - (\mathrm{tr}(\hat{\Sigma}))^2/(pn)$.

We run all the following algorithms over the non-increasing sequence $\lambda_t = \lambda_{\max}10^{-\delta(t-1)/(T-1)}, t \in [T]$ with the default value $\delta = 2, T = 100$. The regularization grid for the joint estimations (Scaled-Lasso, with solver from [25] (SZ), Smoothed Concomitant Lasso (SC), Square-root Lasso [5] (SQRT-Lasso) and the estimator introduced in [23] (SBvG)) begins at $\lambda_{\max}$ given in Proposition 3 with the default value $\sigma_0 = \|y\|/\sqrt{n} \times 10^{-2}$ whereas the grid for the Lasso (L) estimators begins with $\lambda_{\max}^L = \|X^\top y\|_\infty/n$. The Lasso with the universal parameter $\lambda = \sqrt{2\log(p)/n}$ is denoted (L_U) and SZ refers to Concomitant Lasso with the quantile regularization described in [26] in Fig. 1.

For each method, 50 replications are computed from the model aforementioned. Results are presented as boxplots in Fig. 1 (see Appendix E for additional settings).

## 4.1 Performance standard deviation estimators

As noted earlier in [12], spurious correlations can strongly affect sparse regression and usually lead to large biases. This makes the standard deviation estimation very challenging and affects the cross-validation estimator based on the Lasso as they usually underestimate the standard deviation. The phenomenon is amplified when one uses least squares refitting on the cross-validated Lasso, as noticed in [22]. Here we show an example where refitting cross-validation degrades the estimation.

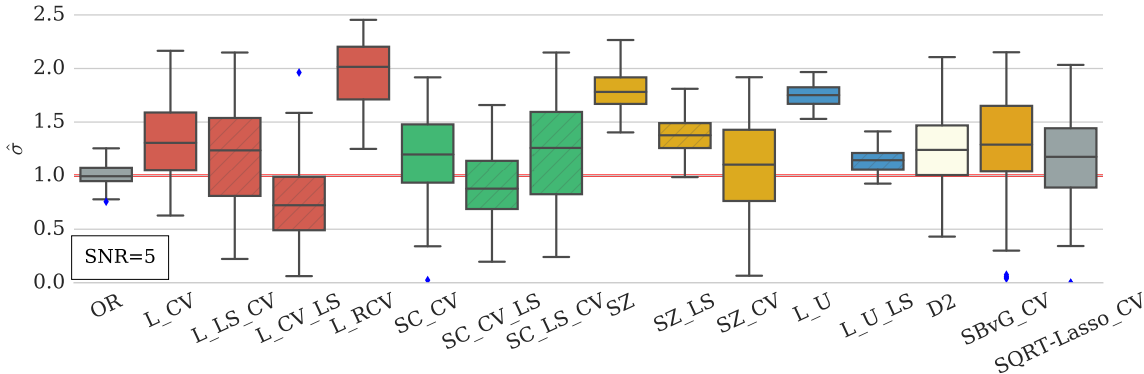

Figure 1: Estimation performance on synthetic dataset ($n = 100, p = 500, \rho = 0.6, \text{snr} = 5, s = 0.9$).

In our experiments, we observe that SC and SZ are very efficient in high sparsity settings with low correlations, correcting for the positive bias of the estimator estimator from [23] (SBvG). In [22], it was also argued that the cross-validation estimator based on Lasso is more stable and performs better when the sparsity decreases and when the snr increases. We would like to emphasize that this is not as the case when one performs a cross-validation procedure on the Concomitant Lasso. Here, we show that the latter outperforms the Lasso estimators and has a lower variance in most cases, especially when applying least squares refitting.

## 4.2 Computational performance

Figure 2(a) presents on the Leukemia dataset the computation times observed for the different CV methods. The Smoothed Concomitant Lasso is based on the coordinate descent algorithm described in Algorithm 1, written in Python and Cython to generate low level C code, offering high performance. When a Lasso solver is needed, we have used the one from `scikit-learn`, that is coded similarly. For SZ_CV, computations are quite heavy as one uses the alternating algorithm proposed in [25]. Depending on the regularization parameter (for instance when one approaches $\lambda_{\min}$) the SZ_CV method is quite intractable and the algorithm faces the numerical issues mentioned earlier. The generic solver used for SBvG and SQRT-Lasso, is the `CVXPY` package [8], explaining why these methods are two orders of magnitude slower than a Lasso. This is in contrast to our solver that reaches similar computing time w.r.t. an efficient Lasso solver, with the additional benefit of jointly estimating the coefficients and the standard deviation of the noise.

Last but not least, Figure 2(b) shows the benefit one can obtain thanks to the safe screening rules introduced above. The Bound safe rule on the Smoothed Concomitant Lasso problem does not

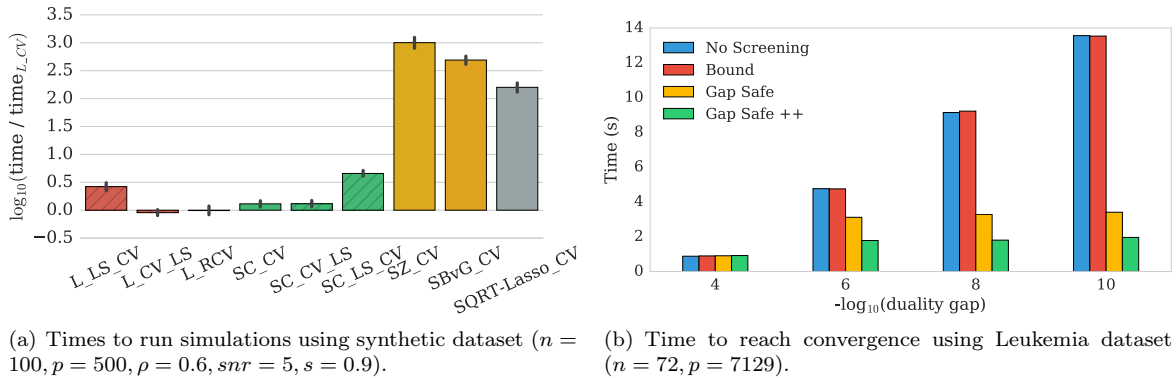

(a) Times to run simulations using synthetic dataset ($n = 100, p = 500, \rho = 0.6, snr = 5, s = 0.9$).

(b) Time to reach convergence using Leukemia dataset ($n = 72, p = 7129$).

Figure 2: Left: comparisons of the computational times using different estimation method (time presented relative to the mean time of the Lasso). Right: speed up using screening rules for the Smoothed Concomitant Lasso w.r.t. to duality gap and for $(\lambda_t)_{t \in [T]}$ with $T = 100$.

show significant acceleration w.r.t. the Gap Safe rule. Indeed, the Gap Safe rule greatly benefits from the convergence of the dual vector, leading to smaller and smaller safe sphere as the iterations proceeds [13, 17]. Another nice feature for the Gap Safe rules relies on a new warm start strategy when computing the full grid $(\lambda_t)_{t \in T}$. For a new $\lambda$, one first performs the optimization over the safe active set (*i.e.,* the non discarded variables) from the previous $\lambda$. This strategy, coined Gap Safe ++, improves the warm start by providing a better primal vector. It helps achieving solutions with great precision at lower cost (up to 8× speed-up on the Leukemia dataset). Pseudo-code and algorithmic details are provided in Appendix A.

## 5 Conclusion

We have explored the joint estimation of the coefficients and of the noise level for $\ell_1$ regularized regression. We have corrected some numerical drawbacks of the Concomitant Lasso estimator by proposing a slightly smoother formulation, leading to the Smoothed Concomitant Lasso. A fast algorithm, relying both on coordinate descent and on safe screening rules with improved warm start was investigated, and it was shown to achieve the same numerical efficiency for the Smoothed Concomitant Lasso as for the Lasso. We also have illustrated on experiments that combined with cross-validation and least-square refitting, Smoothed Concomitant Lasso can outperform the estimators based on the standard Lasso in various settings. It would be interesting in future research to extend our work to more general data-fitting terms [19] and to combine sketching techniques as in [21].

## References

[1] A. Antoniadis. Comments on: $\ell_1$-penalization for mixture regression models. *TEST*, 19(2):257–258, 2010.

[2] H. H. Bauschke and P. L. Combettes. *Convex analysis and monotone operator theory in Hilbert spaces*. Springer, New York, 2011.

[3] S. R. Becker, E. J. Candès, and M. C. Grant. Templates for convex cone problems with applications to sparse signal recovery. *Math. Program. Comput.*, 3(3):165–218, 2011.

[4] A. Belloni and V. Chernozhukov. Least squares after model selection in high-dimensional sparse models. *Bernoulli*, 19(2):521–547, 2013.

[5] A. Belloni, V. Chernozhukov, and L. Wang. Square-root Lasso: Pivotal recovery of sparse signals via conic programming. *Biometrika*, 98(4):791–806, 2011.

[6] P. J. Bickel, Y. Ritov, and A. B. Tsybakov. Simultaneous analysis of Lasso and Dantzig selector. *Ann. Statist.*, 37(4):1705–1732, 2009.

[7] S. Chrétien and S. Darses. Sparse recovery with unknown variance: a lasso-type approach. *IEEE Trans. Inf. Theory*, 2011.

[8] Steven Diamond and Stephen Boyd. CVXPY: A Python-embedded modeling language for convex optimization. *Journal of Machine Learning Research*, 2016. To appear.

[9] L. Dicker. Variance estimation in high-dimensional linear models. *Biometrika*, 101(2):269–284, 2014.

[10] L. El Ghaoui, V. Viallon, and T. Rabbani. Safe feature elimination in sparse supervised learning. *J. Pacific Optim.*, 8(4):667–698, 2012.

[11] F. Pedregosa *et al.* Scikit-learn: Machine learning in Python. *J. Mach. Learn. Res.*, 12:2825–2830, 2011.

[12] J. Fan, S. Guo, and N. Hao. Variance estimation using refitted cross-validation in ultrahigh dimensional regression. *J. Roy. Statist. Soc. Ser. B*, 74(1):37–65, 2012.

[13] O. Fercoq, A. Gramfort, and J. Salmon. Mind the duality gap: safer rules for the lasso. In *ICML*, pages 333–342, 2015.

[14] J. Friedman, T. Hastie, H. Höfling, and R. Tibshirani. Pathwise coordinate optimization. *Ann. Appl. Stat.*, 1(2):302–332, 2007.

[15] P. J. Huber. *Robust Statistics*. John Wiley & Sons Inc., 1981.

[16] J. Lederer. Trust, but verify: benefits and pitfalls of least-squares refitting in high dimensions. *arXiv preprint arXiv:1306.0113*, 2013.

[17] E. Ndiaye, O. Fercoq, A. Gramfort, and J. Salmon. Gap safe screening rules for sparse multi-task and multi-class models. In *NIPS*, pages 811–819, 2015.

[18] Y. Nesterov. Smooth minimization of non-smooth functions. *Math. Program.*, 103(1):127–152, 2005.

[19] A. B. Owen. A robust hybrid of lasso and ridge regression. *Contemporary Mathematics*, 443:59–72, 2007.

[20] J. Peypouquet. *Convex optimization in normed spaces: theory, methods and examples.* Springer, 2015.

[21] V. Pham and L. El Ghaoui. Robust sketching for multiple square-root LASSO problems. In *AISTATS*, 2015.

[22] S. Reid, R. Tibshirani, and J. Friedman. A study of error variance estimation in lasso regression. *arXiv preprint arXiv:1311.5274*, 2013.

[23] N. Städler, P. Bühlmann, and S. van de Geer. $\ell_1$-penalization for mixture regression models. *TEST*, 19(2):209–256, 2010.

[24] T. Sun and C.-H. Zhang. Comments on: $\ell_1$-penalization for mixture regression models. *TEST*, 19(2):270–275, 2010.

[25] T. Sun and C.-H. Zhang. Scaled sparse linear regression. *Biometrika*, 99(4):879–898, 2012.

[26] T. Sun and C.-H. Zhang. Sparse matrix inversion with scaled lasso. *Journal of Machine Learning Research*, 14:3385–3418, 2013.

[27] R. Tibshirani. Regression shrinkage and selection via the lasso. *J. Roy. Statist. Soc. Ser. B*, 58(1):267–288, 1996.

[28] H. Xu, C. Caramanis, and S. Mannor. Robust regression and lasso. *IEEE Trans. Inf. Theory*, 56(7):3561–3574, 2010.

[29] S. Yun. On the iteration complexity of cyclic coordinate gradient descent methods. *SIAM Journal on Optimization*, 24(3):1567–1580, 2014.

# A Algorithms

## A.1 Scaled-Lasso Algorithm (SZ)

We describe in Algorithm 2 the algorithm proposed by Sun and Zhang in [25] to compute the Scaled-Lasso.

In our experiments we have used it for computing the SZ results, using the default parameters of the associated R packages `scalreg-package`, and a choice of $\lambda$'s following the quantile oriented one described in [26].

Note that contrary to our approach the stopping criterion is only based on checking the absence of consecutive increments on the noise level, whereas we consider dual gap evaluations as a more principle way.

Concerning the Lasso steps, as for other Lasso computations in our experiments, we have used the Lasso solver from `scikit-learn` with a dual gap tolerance of $10^{-4}$ and the other parameters set to their default values.

---

**Algorithm 2:** Scaled-Lasso algorithm [25] for a fixed $\lambda$ value

**Input**   : $X, y, \epsilon(=10^{-4}), K=100, \lambda, \sigma_{\text{old}}(=5), \sigma_{\text{new}}(=0.1)$
$k = 0$
**while** $|\sigma_{\text{old}} - \sigma_{\text{new}}| > \epsilon$ *and* $k < K$ **do**
  $k \leftarrow k + 1$
  $\sigma_{\text{old}} = \sigma_{\text{new}}$
  $\lambda^L \leftarrow \lambda \sigma_{\text{old}}$
  $\beta \leftarrow \underset{\beta \in \mathbb{R}^p}{\arg\min} \; \frac{1}{2n} \|y - X\beta\|^2 + \lambda^L \|\beta\|_1$       // Lasso step with parameter $\lambda^L$
  $\sigma_{\text{new}} = \|y - X\beta\| / \sqrt{n}$         // Noise estimation step
**Output**: $(\beta, \sigma_{\text{new}})$

---

## A.2 Smoothed Concomitant Lasso algorithm (SC)

We first present the inner loop of our main algorithm, *i.e.*, the implementation of coordinate descent for the Smoothed Concomitant Lasso. In Algorithm 3, we denote by $\mathcal{A}$ the active set, *i.e.*, the set of coordinates that we have not screened out. For safe screening rules, this set is guaranteed to contain the support of the optimal solution.

---

**Algorithm 3:** CD4SCL – Coordinate Descent for the Smoothed Concomitant Lasso with Gap Safe screening

---

**Input** : $X, y, \epsilon, K, f^{\text{ce}}(=10), \lambda, \sigma_0, \beta, \sigma$
$\mathcal{A} \leftarrow [p]$
**for** $k \in [K]$ **do**
    **if** $k \mod f^{\text{ce}} = 1$ **then**
        Compute $\theta$ thanks to (9)
        **if** $G_{\lambda,\sigma_0}(\beta, \sigma, \theta) = P_{\lambda_t,\sigma_0}(\beta, \sigma) - D_{\lambda_t}(\theta) \leq \epsilon$ **then**   // Stopping criterion
        | **break**
        Update $\mathcal{A}$ thanks to Theorem 6   // Screening test
    **for** $j \in \mathcal{A}$ **do**   // Loop over coordinates
        $\beta_j \leftarrow \mathcal{S}_{n\sigma\lambda_t/\|X_j\|^2}\left(\beta_j - X_j^\top(X\beta - y)/\|X_j\|^2\right)$   // Soft-thresholding step
        $\sigma \leftarrow \sigma_0 \vee (\|y - X\beta\|/\sqrt{n}))$   // Noise estimation step
**Output**: $\beta, \sigma, \mathcal{A}$

---

We now present in Algorithm 4, the fast solver we proposed for the Smoothed Concomitant Lasso, relying on the three following key features: coordinate descent, Gap Safe screening rules and improved warm start propositions.

In Algorithm 4, the first occurrence of CD4SCL is a warming step aimed at improving the current primal point at a low cost. For Gap Safe, we disable it by setting $K_0 = 0$. For the experiments with Gap Safe ++, we have set $K_0 = K = 5000$ and $\epsilon_0 = \epsilon$. Note that in the first inner loop, we restrict the design matrix to the coordinates in $\mathcal{A}$, the previous active set, which is not safe any more. The duality gap and other quantities need to be modified accordingly.

Concerning the parameter $f^{\text{ce}}$ it governs how often we perform the dual gap evaluation. Due to the complexity of this step, we do not recommend to do this step every pass over the features, but rather compute this quantity less often, every $f^{\text{ce}}$ passes. In practice we have fixed its value to $f^{\text{ce}} = 10$ for all our experiments.

---

**Algorithm 4:** Coordinate Descent for the Smoothed Concomitant Lasso with Gap Safe ++ screening

---

**Input** : $X, y, \epsilon, \epsilon_0, K, K_0, f^{\text{ce}}, (\lambda_t)_{t \in [T-1]}, \sigma_0$
$\lambda_0 = \lambda_{\max} = \|X^\top y\|_\infty/(\|y\|\sqrt{n}), \quad \beta^{\lambda_0} = 0, \quad \sigma^{\lambda_0} = \|y\|/\sqrt{n}$
$\mathcal{A} \leftarrow [p]$
**for** $t \in [T-1]$ **do**
    $\beta, \sigma \leftarrow \beta^{\lambda_{t-1}}, \sigma^{\lambda_{t-1}}$ (previous $\epsilon$-solution)   // Get previous $\epsilon$-solution
    $\beta, \sigma, \_ \leftarrow \text{CD4SCL}(X_\mathcal{A}, y, \epsilon_0, K_0, f^{\text{ce}}, \lambda_t, \sigma_0, \beta, \sigma)$   // Warm start ++ step
    $\beta, \sigma, \mathcal{A} \leftarrow \text{CD4SCL}(X, y, \epsilon, K, f^{\text{ce}}, \lambda_t, \sigma_0, \beta, \sigma)$   // Standard loop
    $\beta^{\lambda_t}, \sigma^{\lambda_t} \leftarrow \beta, \sigma$
**Output**: $(\beta^{\lambda_t})_{t \in [T-1]}, (\sigma^{\lambda_t})_{t \in [T-1]}$

---

# B Experiments

In this section, we present some extensive benchmarks with the synthetic dataset with less sparse signal than in Section 4. The main observation is that Smoothed Concomitant Lasso with cross-validation is stable w.r.t. various settings and leads to better performance than other Lasso variants.

For each setting, we compare the mean running time for 50 simulations. The results are displayed in Fig. 4. The computational time of our algorithm is in the same order than the usual one for the Lasso.

## C  Link with the perspective of a function

The concomitant scale estimator introduced by Huber [15, Ch. 7.7 and 7.8] (see also [19, 1]), is related to the perspective of a function defined for a convex function $f : \mathbb{R}^n \to \mathbb{R} \cup \{+\infty\}$ as the function $\text{persp}_f : \mathbb{R}^n \times \mathbb{R} \to \mathbb{R} \cup \{+\infty\}$ such that

$$\text{persp}_f(r, \sigma) = \begin{cases} \sigma f\left(\frac{r}{\sigma}\right), & \text{if } \sigma > 0, \\ +\infty, & \text{if } \sigma \leq 0. \end{cases}$$

This function is not lower semi-continuous in general. However, lower semi-continuity is a very desirable property. Together with the fact that the function is infinite at infinity, this guarantees the existence of minimizers [20, Theorem 2.19]. Hence we consider instead its biconjugate, which is always lower semi-continuous [2, Theorem 13.32]. One can show [2, Example 13.8] that the Fenchel conjugate of $\text{persp}_f$ is

$$\text{persp}_f^*(\theta, \nu) = \begin{cases} 0, & \text{if } \nu + f^*(\theta) \leq 0, \\ +\infty, & \text{otherwise.} \end{cases}$$

Hence a direct calculation shows that

**Proposition 8.**

$$\text{persp}_f^{**}(r, \sigma) = \begin{cases} \sigma f^{**}\left(\frac{r}{\sigma}\right), & \text{if } \sigma > 0, \\ \sup_{\theta \in \text{dom } f^*} \langle \theta, r \rangle, & \text{if } \sigma = 0, \\ +\infty & \text{otherwise.} \end{cases}$$

*Proof.* Let us define $g = \text{persp}_f^*$ for simplicity.

<u>First case</u>: $\sigma > 0$.

$$\text{persp}_f^{**}(r, \sigma) = \sup_{\theta \in \mathbb{R}^n, \nu \in \mathbb{R}} \langle \theta, r \rangle + \sigma \nu - g(\theta, \nu) = \sup_{\theta \in \mathbb{R}^n, \nu \in \mathbb{R}} \{\langle \theta, r \rangle + \sigma \nu : \nu + f^*(\theta) \leq 0\}$$

As $\sigma > 0$, for a given $\beta$, one should take $\nu$ the largest possible, hence $\nu = -f^*(\theta)$.

$$\text{persp}_f^{**}(r, \sigma) = \sup_{\theta \in \mathbb{R}^n} \langle \theta, r \rangle - \sigma f^*(\theta) = \sigma \sup_{\theta \in \mathbb{R}^n} \langle \theta, r/\sigma \rangle - f^*(\theta) = \sigma f^{**}(r/\sigma)$$

<u>Second case</u>: $\sigma = 0$.

$$\text{persp}_f^{**}(r, 0) = \sup_{\theta \in \mathbb{R}^n, \nu \in \mathbb{R}} \langle \theta, r \rangle - g(\theta, \nu) = \sup_{\theta \in \mathbb{R}^n, \nu \in \mathbb{R}} \{\langle \theta, r \rangle : \nu + f^*(\theta) \leq 0\}.$$

As $\nu$ has no influence on the value of the objective, we can choose it as small as we want and so the only requirement on $\theta$ is that it should belong to the domain of $f^*$. We get

$$\text{persp}_f^{**}(r, 0) = \sup_{\theta \in \text{dom } f^*} \langle \theta, r \rangle$$

<u>Third case</u>: $\sigma < 0$. If $\sigma < 0$, we can let $\nu$ go to $-\infty$ in the formula of $\text{persp}_f^{**}(r,\sigma)$ which leads to $\text{persp}_f^{**}(r,\sigma) = +\infty$. □

In our case, $f(r) = \frac{1}{2n}\|r\|_2^2 + \frac{1}{2}$ and so $f^{**} = f$ and $\text{dom } f^* = \mathbb{R}^n$. Hence, we get

$$\text{persp}_f^{**}(r,\sigma) = \begin{cases} \frac{1}{2n\sigma}\|r\|_2^2 + \frac{\sigma}{2}, & \text{if } \sigma > 0, \\ 0, & \text{if } \sigma = 0 \text{ and } r = 0, \\ +\infty, & \text{otherwise.} \end{cases}$$

Taking this lower semi-continuous function leads to a well defined Concomitant Lasso estimator thanks to the following formulation

$$(\hat{\beta}^{(\lambda)}, \hat{\sigma}^{(\lambda)}) \in \underset{\beta \in \mathbb{R}^p, \sigma \in \mathbb{R}}{\arg\min} \ \text{persp}_f^{**}(y - X\beta, \sigma) + \lambda \|\beta\|_1.$$

The only difference with the original one is that we take $\hat{\sigma}^{(\lambda)} = 0$ if $y - X\hat{\beta}^{(\lambda)} = 0$.

## D  Dual of the Smoothed Concomitant Lasso

**Theorem 2.** *For $\lambda > 0$ and $\sigma_0 > 0$, the Smoothed Concomitant Lasso estimator $\hat{\beta}^{(\lambda,\sigma_0)}$ and its associated noise level estimate $\hat{\sigma}^{(\lambda,\sigma_0)}$ are defined as solutions of the primal optimization problem*

$$(\hat{\beta}^{(\lambda,\sigma_0)}, \hat{\sigma}^{(\lambda,\sigma_0)}) \in \underset{\beta \in \mathbb{R}^p, \sigma \geq \sigma_0}{\arg\min} \frac{1}{2n\sigma}\|y - X\beta\|^2 + \frac{\sigma}{2} + \lambda \|\beta\|_1, \tag{12}$$

*With $\Delta_{X,\lambda} = \{\theta \in \mathbb{R}^n : \|X^\top \theta\|_\infty \leq 1, \|\theta\| \leq 1/(\lambda\sqrt{n})\}$, the dual formulation of the Smoothed Concomitant Lasso reads*

$$\hat{\theta}^{(\lambda,\sigma_0)} = \underset{\theta \in \Delta_{X,\lambda}}{\arg\max} \underbrace{\langle y, \lambda\theta \rangle + \sigma_0 \left(\frac{1}{2} - \frac{\lambda^2 n}{2}\|\theta\|^2\right)}_{D_{\lambda,\sigma_0}(\theta)}. \tag{13}$$

*For an optimal primal vector $\hat{\beta}^{(\lambda,\sigma_0)}$, we must have $\hat{\sigma}^{(\lambda,\sigma_0)} = \sigma_0 \vee (\|y - X\hat{\beta}^{(\lambda,\sigma_0)}\|/\sqrt{n})$. We also have the link-equation between primal and dual solutions: $y = n\lambda\hat{\sigma}^{(\lambda,\sigma_0)}\hat{\theta}^{(\lambda,\sigma_0)} + X\hat{\beta}^{(\lambda,\sigma_0)}$.*

*Proof.*

$$\min_{\beta \in \mathbb{R}^p, \sigma \geq \sigma_0} \frac{1}{2n\sigma}\|y - X\beta\|^2 + \frac{\sigma}{2} + \lambda \|\beta\|_1$$

$$= \min_{\beta \in \mathbb{R}^p, z \in \mathbb{R}^n, \sigma \geq \sigma_0} \frac{1}{2n\sigma}\|y - z\|^2 + \frac{\sigma}{2} + \lambda \|\beta\|_1 \text{ s.t. } z = X\beta$$

$$= \min_{\beta \in \mathbb{R}^p, z \in \mathbb{R}^n, \sigma \geq \sigma_0} \max_{\theta \in \mathbb{R}^n} \underbrace{\frac{1}{2n\sigma}\|y - z\|^2 + \frac{\sigma}{2} + \lambda \|\beta\|_1 + \lambda\theta^\top(z - X\beta)}_{\mathcal{L}(\beta,\sigma,\theta,z)},$$

$$= \max_{\theta \in \mathbb{R}^n} \min_{\sigma \geq \sigma_0} \frac{\sigma}{2} - \max_{z \in \mathbb{R}^n}\left\{\langle -\lambda\theta, z\rangle - \frac{1}{2n\sigma}\|y - z\|^2\right\} - \lambda \max_{\beta \in \mathbb{R}^p}\langle X^\top\theta, \beta\rangle - \|\beta\|_1,$$

$$= \max_{\theta \in \mathbb{R}^n} \min_{\sigma \geq \sigma_0} \frac{\sigma}{2} - \frac{\lambda^2 n\sigma}{2}\|\theta\|^2 + \langle \lambda(\theta, y) - \iota_{\mathcal{B}_\infty}(X^\top\theta)\rangle.$$

The fourth line is true because the Slater's condition is met, hence we can permute min and max thanks to strong duality. Finally we obtain the dual problem since

$$\min_{\sigma \geq \sigma_0} \sigma \left( \frac{1}{2} - \frac{\lambda^2 n}{2} \|\theta\|^2 \right) = \begin{cases} \sigma_0 \left( \frac{1}{2} - \frac{\lambda^2 n}{2} \|\theta\|^2 \right), & \text{if } \frac{1}{2} - \frac{\lambda^2 n}{2} \|\theta\|^2 \geq 0, \\ -\infty, & \text{otherwise.} \end{cases}$$

□

Let us denote

$$\left( \hat{\beta}^{(\lambda,\sigma_0)}, \hat{\sigma}^{(\lambda,\sigma_0)}, \hat{\theta}^{(\lambda,\sigma_0)}, \hat{z}^{(\lambda,\sigma_0)} \right) \in \underset{\beta \in \mathbb{R}^p, z \in \mathbb{R}^n, \sigma \geq \sigma_0}{\arg\min} \max_{\theta \in \mathbb{R}^n} \mathcal{L}(\beta, \sigma, \theta, z).$$

The primal-dual link equation follows directly from the Fermat's rule:

$$\frac{\partial \mathcal{L}(\hat{\beta}^{(\lambda,\sigma_0)}, \hat{\sigma}^{(\lambda,\sigma_0)}, \cdot, \hat{z}^{(\lambda,\sigma_0)})}{\partial \theta}(\hat{\theta}^{(\lambda,\sigma_0)}) = \hat{z}^{(\lambda,\sigma_0)} - X\hat{\beta}^{(\lambda,\sigma_0)} = 0,$$

$$\frac{\partial \mathcal{L}(\hat{\beta}^{(\lambda,\sigma_0)}, \hat{\sigma}^{(\lambda,\sigma_0)}, \hat{\theta}^{(\lambda,\sigma_0)}, \cdot)}{\partial z}(\hat{z}^{(\lambda,\sigma_0)}) = -\frac{1}{n\hat{\sigma}^{(\lambda,\sigma_0)}}(y - \hat{z}^{(\lambda,\sigma_0)}) + \lambda \hat{\theta}^{(\lambda,\sigma_0)} = 0.$$

# E  Convergence

**Proposition 4.** *Let $(\beta_k)_{k \in \mathbb{N}}$ be a sequence that converges to $\hat{\beta}^{(\lambda,\sigma_0)}$. Then $(\theta_k)_{k \in \mathbb{N}}$ built from $\theta_k = (y - X\beta_k)/((\lambda n \sigma_0) \vee \|X^\top(y - X\beta_k)\|_\infty \vee (\lambda\sqrt{n}\|y - X\beta_k\|))$ converges to $\hat{\theta}^{(\lambda,\sigma_0)}$. Hence the sequence of dual gap $(G_{\lambda,\sigma_0}(\beta_k, \sigma_k, \theta_k))_{k \in \mathbb{N}}$ converges to zero.*

*Proof.* Let $\alpha_k = (\lambda n \sigma_0) \vee (\|X^\top(y - X\beta_k)\|_\infty) \vee (\lambda\sqrt{n}\|y - X\beta_k\|)$, then we have:

$$\left\| \theta_k - \hat{\theta}^{(\lambda,\sigma_0)} \right\| = \left\| \frac{1}{\alpha_k}(y - X\beta_k) - \frac{1}{\lambda n \hat{\sigma}^{(\lambda,\sigma_0)}}(y - X\hat{\beta}^{(\lambda,\sigma_0)}) \right\|$$

$$= \left\| \left( \frac{1}{\alpha_k} - \frac{1}{\lambda n \hat{\sigma}^{(\lambda,\sigma_0)}} \right)(y - X\beta_k) - \frac{(X\hat{\beta}^{(\lambda,\sigma_0)} - X\beta_k)}{\lambda n \hat{\sigma}^{(\lambda,\sigma_0)}} \right\|$$

$$\leq \left| \frac{1}{\alpha_k} - \frac{1}{\lambda n \hat{\sigma}^{(\lambda,\sigma_0)}} \right| \|y - X\beta_k\| + \left\| \frac{X\hat{\beta}^{(\lambda,\sigma_0)} - X\beta_k}{\lambda} \right\|.$$

If $\beta_k \to \hat{\beta}^{(\lambda,\sigma_0)}$, then the second term in the last display converges to zero, and for the first term, we show below that $\alpha_k \to \alpha := (\lambda n \sigma_0) \vee (\|X^\top(y - X\hat{\beta}^{(\lambda,\sigma_0)})\|_\infty) \vee (\lambda\sqrt{n}\|y - X\hat{\beta}^{(\lambda,\sigma_0)}\|)$. Recall that from Fermat's rule, we have $y - X\hat{\beta}^{(\lambda,\sigma_0)} = \lambda n \hat{\sigma}^{(\lambda,\sigma_0)} \hat{\theta}^{(\lambda,\sigma_0)}$ and $X^\top(y - X\hat{\beta}^{(\lambda,\sigma_0)}) \in \lambda n \hat{\sigma}^{(\lambda,\sigma_0)} \partial \|\cdot\|_1(\hat{\beta}^{(\lambda,\sigma_0)})$, leading to one of the three following situations:

- if $\hat{\sigma}^{(\lambda,\sigma_0)} > \sigma_0$, then $\|X^\top(y - X\hat{\beta}^{(\lambda,\sigma_0)})\|_\infty \leq \lambda n \hat{\sigma}^{(\lambda,\sigma_0)} = \lambda\sqrt{n}\|y - X\hat{\beta}^{(\lambda,\sigma_0)}\| = \alpha$.

- If $\hat{\sigma}^{(\lambda,\sigma_0)} = \sigma_0$ and $\hat{\beta}^{(\lambda,\sigma_0)} \neq 0$, we have $X^\top(y - X\hat{\beta}^{(\lambda,\sigma_0)}) = \lambda n \hat{\sigma}^{(\lambda,\sigma_0)} \hat{v}$ where $\hat{v} \in \partial \|\cdot\|_1(\hat{\beta}^{(\lambda,\sigma_0)})$. Since $\hat{\beta}^{(\lambda,\sigma_0)} \neq 0$, there exists a coordinate $j$ such that $\hat{\beta}_j^{(\lambda,\sigma_0)} \neq 0$ and so $|\hat{v}_j| = 1$ which implies that $\|\hat{v}\|_\infty = 1$. Hence $\|X^\top(y - X\hat{\beta}^{(\lambda,\sigma_0)})\|_\infty = \lambda n \hat{\sigma}^{(\lambda,\sigma_0)}$. Moreover, $\|y - X\hat{\beta}^{(\lambda,\sigma_0)}\| = \lambda n \hat{\sigma}^{(\lambda,\sigma_0)} \|\hat{\theta}^{(\lambda,\sigma_0)}\| \leq \lambda n \hat{\sigma}^{(\lambda,\sigma_0)}/(\lambda\sqrt{n})$ since $\hat{\theta}^{(\lambda,\sigma_0)} \in \Delta_{X,\lambda}$. Hence, $\lambda\sqrt{n}\|y - X\hat{\beta}^{(\lambda,\sigma_0)}\| \leq \lambda n \hat{\sigma}^{(\lambda,\sigma_0)} = \|X^\top(y - X\hat{\beta}^{(\lambda,\sigma_0)})\|_\infty = \alpha$.

- If $\hat{\sigma}^{(\lambda,\sigma_0)} = \sigma_0$ and $\hat{\beta}^{(\lambda,\sigma_0)} = 0$, then $y = \lambda n \sigma_0 \hat{\theta}^{(\lambda,\sigma_0)}$, $\lambda\sqrt{n}\|y\| \leq \lambda n \sigma_0$ since $\hat{\theta}^{(\lambda,\sigma_0)} \in \Delta_{X,\lambda}$, and $\|X^\top y\|_\infty \leq \lambda n \sigma_0$. Hence $\alpha = \lambda n \sigma_0$.

  Finally, we have shown that in all cases, $(\alpha_k)_{k\in\mathbb{N}}$ converges to $\alpha = \lambda n \hat{\sigma}^{(\lambda,\sigma_0)}$, so the first term also converges to zero. $\square$

**Proposition 5.** $\forall \beta \in \mathbb{R}^p, \theta \in \Delta_{X,\lambda}, \sigma \geq \sigma_0, \quad G^{\mathrm{L}}_{\sigma\lambda}(\beta, \theta) \leq \sigma G_{\lambda,\sigma_0}(\beta, \sigma, \theta)$.

*Proof.* Since $\sigma - \sigma_0 \geq 0$ and $\lambda\sqrt{n}\|\theta\| \leq 1$, we have

$$\frac{\sigma\lambda^2 n}{2}\|\theta\|^2 = \frac{\sigma-\sigma_0}{2}\lambda^2 n\|\theta\|^2 + \frac{\sigma_0\lambda^2 n}{2}\|\theta\|^2 \leq \frac{\sigma-\sigma_0}{2} + \frac{\sigma_0\lambda^2 n}{2}\|\theta\|^2.$$

$$\begin{aligned}
G^{\mathrm{L}}_{\sigma\lambda}(\beta, \theta) &= P^{\mathrm{L}}_{\sigma\lambda}(\beta) - D^{\mathrm{L}}_{\sigma\lambda}(\theta) \\
&= \frac{1}{2n}\|y - X\beta\|^2 + \sigma\lambda\|\beta\|_1 - \frac{1}{2n}\|y\|^2 + \frac{1}{2n}\|y - \sigma\lambda n\theta\|^2 \\
&= \frac{1}{2n}\|y - X\beta\|^2 + \sigma\lambda\|\beta\|_1 - \sigma\lambda\langle y, \theta\rangle + \frac{\sigma^2\lambda^2 n}{2}\|\theta\|^2 \\
&\leq \sigma\left(\frac{1}{2n\sigma}\|y - X\beta\|^2 + \lambda\|\beta\|_1 - \lambda\langle y, \theta\rangle + \frac{\sigma}{2} - \sigma_0(\frac{1}{2} - \frac{\lambda^2 n}{2}\|\theta\|^2)\right) \\
&= \sigma\left(P_{\lambda,\sigma_0}(\beta, \sigma) - D_{\lambda,\sigma_0}(\theta)\right) = \sigma G_{\lambda,\sigma_0}(\beta, \theta). \quad \square
\end{aligned}$$

# F  Safe Rules

**Proposition 6.** *For all* $(\beta, \sigma, \theta) \in \mathbb{R}^p \times \mathbb{R}_+ \times \Delta_{X,\lambda}$, *then for*

$$r = \sqrt{\frac{2G_{\lambda,\sigma_0}(\beta, \sigma, \theta)}{\lambda^2 \sigma_0 n}},$$

*we have* $\hat{\theta}^{(\lambda,\sigma_0)} \in \mathcal{B}(\theta, r)$. *Thus, we have the following safe sphere screening rule*

$$|X_j^\top \theta| + r\|X_j\| < 1 \implies \hat{\beta}^{(\lambda,\sigma_0)}_j = 0. \tag{14}$$

*Proof.* The proof follows [17]. We give it for the sake of completeness. By weak duality, $\forall \beta \in \mathbb{R}^p, D_{\lambda,\sigma_0}(\theta) \leq P_\lambda(\beta, \sigma)$. Then, note that the dual objective function of the Smoothed Concomitant Lasso is $\lambda^2 \sigma_0 n$-strongly concave. This implies:

$$\forall (\theta, \theta') \in \Delta_{X,\lambda} \times \Delta_{X,\lambda}, \quad D_{\lambda,\sigma_0}(\theta) \leq D_{\lambda,\sigma_0}(\theta') + \nabla D_{\lambda,\sigma_0}(\theta')^\top(\theta - \theta') - \frac{\lambda^2 \sigma_0 n}{2}\|\theta - \theta'\|^2.$$

Moreover, since $\hat{\theta}^{(\lambda,\sigma_0)}$ maximizes the concave function $D_{\lambda,\sigma_0}$, the following inequality holds true:

$$\forall \theta \in \Delta_{X,\lambda}, \quad \nabla D_\lambda(\hat{\theta}^{(\lambda,\sigma_0)})^\top(\theta - \hat{\theta}^{(\lambda,\sigma_0)}) \leq 0.$$

Hence, we have for all $\theta \in \Delta_{X,\lambda}$ and $\beta \in \mathbb{R}^p$:

$$\begin{aligned}
\frac{\lambda^2 \sigma_0 n}{2}\left\|\theta - \hat{\theta}^{(\lambda,\sigma_0)}\right\|^2 &\leq D_{\lambda,\sigma_0}(\hat{\theta}^{(\lambda,\sigma_0)}) - D_\lambda(\theta) \\
&\leq P_\lambda(\beta, \sigma) - D_{\lambda,\sigma_0}(\theta) = G_{\lambda,\sigma_0}(\beta, \sigma, \theta).
\end{aligned}$$

Furthermore,

$$\max_{\bar{\theta}\in\mathcal{B}(\theta,r)} |X_j^\top \bar{\theta}| \leq |X_j^\top \theta| + \max_{\bar{\theta}\in\mathcal{B}(\theta,r)} |X_j^\top (\bar{\theta}-\theta)| \leq |X_j^\top \theta| + \|X_j\| \max_{\bar{\theta}\in\mathcal{B}(\theta,r)} \|\bar{\theta}-\theta\| = |X_j^\top \theta| + r\,\|X_j\|.$$

Hence $\max_{\bar{\theta}\in\mathcal{B}(\theta,r)} |X_j^\top \bar{\theta}| = |X_j^\top \theta| + r\,\|X_j\|$ since the vector $\bar{\bar{\theta}} := \theta + X_j \frac{r}{\|X_j\|}$ is feasible and attains the bound. $\square$

In this section we derive the Bound Safe screening rules of Proposition 7. First, we need two technical lemmas.

**Lemma 3.** *Let $y'$ and $x$ be two unit vectors, and consider $0 \leq \underline{\gamma} \leq \overline{\gamma} \leq 1$. The optimal value of*

$$\max\{\theta^\top x \quad : \quad \|\theta\| \leq 1, \quad \underline{\gamma} \leq y'^\top \theta \leq \overline{\gamma}\},$$

*is given by*

$$\begin{cases} \overline{\gamma} x^\top y' + \sqrt{1-\overline{\gamma}^2}\sqrt{1-(x^\top y')^2}, & \text{if } x^\top y' > \overline{\gamma}, \\ 1, & \text{if } \underline{\gamma} \leq x^\top y' \leq \overline{\gamma}, \\ \underline{\gamma} x^\top y' + \sqrt{1-\underline{\gamma}^2}\sqrt{1-(x^\top y')^2}, & \text{if } x^\top y' < \underline{\gamma}. \end{cases}$$

*Proof.* First remark that $x$ and $y$ are two privileged directions in the optimization problem at stake. Indeed, if $\theta$ has a nonzero component in a direction orthogonal to both $x$ and $y'$, then, because of the constraint $\|\theta\| = 1$, this reduces the freedom in $\mathrm{Span}(x, y')$ while giving no progress in the objective and the linear constraints. Hence, from now on we can restrict ourselves to the plane $\mathrm{Span}(x, y)$.

We denote by $\angle(w, z) \in \mathbb{R}/2\pi\mathbb{Z}$ the directed angle between unitary vectors $w$ and $z$. We recall that $\cos(\angle(w, z)) = w^\top z$, so we can narrow down our analysis to the three following cases:

a) Assume that $x^\top y' > \overline{\gamma}$. Then the optimal $\theta$ is such that (see Figure (5).(a)) $\|\theta\| = 1$, $\theta^\top y' = \overline{\gamma}$ and $x$ is "between" $\theta$ and $y'$, which implies that $\sin(\angle(\theta, y'))\sin(\angle(y', x)) < 0$. Hence,

$$\begin{aligned} \theta^\top x &= \cos(\angle(\theta, x)) = \cos\left(\angle(\theta, y') + \angle(y', x)\right) \\ &= \cos\left(\angle(\theta, y')\right)\cos\left(\angle(y', x)\right) - \sin\left(\angle(\theta, y')\right)\sin\left(\angle(y', x)\right) \\ &= \theta^\top y'.y'^\top x + |\sin(\angle(\theta, y'))\sin(\angle(y', x))| \\ &= \overline{\gamma} y'^\top x + \sqrt{1-\overline{\gamma}^2}\sqrt{1-(y'^\top x)^2}. \end{aligned}$$

b) Assume that $\underline{\gamma} \leq x^\top y' \leq \overline{\gamma}$, then $\theta = x$ is admissible, and the maximum is 1 (see Figure (5).(b)).

c) Assume that $-1 \leq x^\top y' < \underline{\gamma}$ (see Figure (5).(c)), then the optimal $\theta$ is such that $\|\theta\| = 1$, $\theta^\top y' = \underline{\gamma}$ and $\theta$ is "between" $x$ and $y'$, which implies that $\sin(\angle(\theta, y'))\sin(\angle(y', x)) < 0$. Hence, elementary trigonometry gives

$$\theta^\top x = \cos\left(\angle(\theta, y') + \angle(y', x)\right) = \underline{\gamma} x^\top y' + \sqrt{1-\underline{\gamma}^2}\sqrt{1-(x^\top y')^2}.$$

**Lemma 4.** *Let $y'$ and $x$ be two unit vectors, and consider $0 \leq \underline{\gamma} \leq \overline{\gamma} \leq 1$. The optimal value of*

$$\max\{|\theta^\top x| \quad : \quad \|\theta\| \leq 1, \quad \underline{\gamma} \leq {y'}^\top \theta \leq \overline{\gamma}\},$$

is given by

$$\begin{cases} \overline{\gamma}|x^\top y'| + \sqrt{1-\overline{\gamma}^2}\sqrt{1-(x^\top y')^2}, & \text{if } |x^\top y'| > \overline{\gamma}, \\ 1, & \text{if } \underline{\gamma} \leq |x^\top y'| \leq \overline{\gamma}, \\ \underline{\gamma}|x^\top y'| + \sqrt{1-\underline{\gamma}^2}\sqrt{1-(x^\top y')^2}, & \text{if } |x^\top y'| < \underline{\gamma}. \end{cases}$$

*Proof.* We need to compute

$$\max\{|\theta^\top x| \quad : \quad \|\theta\| \leq 1, \quad \underline{\gamma} \leq {y'}^\top \theta \leq \overline{\gamma}\}.$$

We apply Lemma 3 with $x \leftarrow x$ and $x \leftarrow -x$. We get five cases and for each the value is a maximum between two choices. In fact, one of the two choices is always dominated by the other one. We just present one case for conciseness.

Suppose that $x^\top y' > \overline{\gamma}$ (and thus $-x^\top y' < \underline{\gamma}$ since $\underline{\gamma} \geq 0$). Then the optimal $\theta$ satisfies

$$|\theta^\top x| = \left(\overline{\gamma} x^\top y' + \sqrt{1-\overline{\gamma}^2}\sqrt{1-(x^\top y')^2}\right) \vee \left(-\underline{\gamma} x^\top y' + \sqrt{1-\underline{\gamma}^2}\sqrt{1-(x^\top y')^2}\right).$$

We now remark the equivalence

$$-\underline{\gamma} x^\top y' + \sqrt{1-\underline{\gamma}^2}\sqrt{1-(x^\top y')^2} \leq \overline{\gamma} x^\top y' + \sqrt{1-\overline{\gamma}^2}\sqrt{1-(x^\top y')^2}$$

$$\Leftrightarrow \left(\sqrt{1-\underline{\gamma}^2} - \sqrt{1-\overline{\gamma}^2}\right)\sqrt{1-(x^\top y')^2} - (\overline{\gamma}+\underline{\gamma})x^\top y' \leq 0.$$

This function is decreasing in $x^\top y'$ so

$$\left(\sqrt{1-\underline{\gamma}^2} - \sqrt{1-\overline{\gamma}^2}\right)\sqrt{1-(x^\top y')^2} - (\overline{\gamma}+\underline{\gamma})x^\top y'$$

$$\leq \left(\sqrt{1-\underline{\gamma}^2} - \sqrt{1-\overline{\gamma}^2}\right)\sqrt{1-\overline{\gamma}^2} - (\overline{\gamma}+\underline{\gamma})\overline{\gamma}$$

$$= \sqrt{1-\underline{\gamma}^2}\sqrt{1-\overline{\gamma}^2} - 1 + \overline{\gamma}^2 - \overline{\gamma}^2 - \underline{\gamma}\overline{\gamma} \leq 0.$$

Thus, the second term in the maximum is never selected and we can simplify the expression. The other cases can be handled similarly. □

**Proposition 7.** *Assume that, for a given $\lambda > 0$, we have an upper bound $\overline{\eta} \in (0, +\infty]$, and a lower bound $\underline{\eta} \in (0, +\infty]$ over the Smoothed Concomitant Lasso problem (7). Denote by $x_j = X_j/\|X_j\|$ and $y' = y/\|y\|$ two unit vectors, and by $\underline{\gamma} = (\underline{\eta} - \sigma_0/2)\sqrt{n}/\|y\|$ and $\overline{\gamma} = \overline{\eta}\sqrt{n}/\|y\|$. Then if one of the three following conditions is met*

- $|x_j^\top y'| > \overline{\gamma}$ and $\overline{\gamma}|x_j^\top y'| + \sqrt{1-\overline{\gamma}^2}\sqrt{1-(x_j^\top y')^2} < \lambda\sqrt{n}/\|X_j\|$.

- $\underline{\gamma} \leq |x_j^\top y'| \leq \overline{\gamma}$ and $1 < \lambda\sqrt{n}/\|X_j\|$.
- $|x_j^\top y'| < \underline{\gamma}$ and $\underline{\gamma}|x_j^\top y'| + \sqrt{1-\underline{\gamma}^2}\sqrt{1-(x_j^\top y')^2} < \lambda\sqrt{n}/\|X_j\|$.

*then the $j$-th feature can be discarded* i.e., $\hat{\beta}_j^{(\lambda,\sigma_0)} = 0$.

*Proof.* If $\underline{\eta} \leq D_\lambda(\hat{\theta}^{(\lambda,\sigma_0)})\rangle \leq \overline{\eta}$ and

$$\max\{|X_j^\top \theta| \,:\, \lambda\sqrt{n}\,\|\theta\| \leq 1, \quad \underline{\eta} \leq D_\lambda(\theta)\rangle \leq \overline{\eta}\} < 1, \tag{15}$$

then the $j$-th feature can be discarded (see Eq. (10)).

For the standard Concomitant Lasso formulation, $D_\lambda(\theta) = \langle y, \lambda\theta\rangle$ and Lemma 4 can be directly applied to get a safe screening rule from (15). To treat the Smoothed Concomitant Lasso (7), we check that if $\underline{\eta} \leq D_{\lambda,\sigma_0}(\theta) \leq \overline{\eta}$ then $\underline{\eta} - \sigma_0/2 \leq \langle y, \lambda\theta\rangle \leq \overline{\eta}$. Thus, we obtain a new screening test

$$\max\{|X_j^\top \theta| \,:\, \lambda\sqrt{n}\,\|\theta\| \leq 1, \quad \underline{\eta} - \sigma_0/2 \leq \langle y, \lambda\theta\rangle \leq \overline{\eta}\} < 1. \tag{16}$$

To leverage Lemma 4 we reformulate the test as

$$\max\left\{\left|\frac{\lambda\sqrt{n}}{\|X_j\|}X_j^\top \theta\right| \,:\, \sqrt{n}\lambda\,\|\theta\| \leq 1, \quad \frac{(\underline{\eta}-\sigma_0/2)\sqrt{n}}{\|y\|} \leq \left\langle \frac{y}{\|y\|}, \sqrt{n}\lambda\theta \right\rangle \leq \frac{\overline{\eta}\sqrt{n}}{\|y\|}\right\} < \frac{\sqrt{n}\lambda}{\|X_j\|}.$$

Denoting by $x_j' = X_j/\|X_j\|$ and $y' = y/\|y\|$ two unit vectors, and by $\underline{\gamma} = (\underline{\eta}-\sigma_0/2)\sqrt{n}/\|y\|$ and $\overline{\gamma} = \overline{\eta}\sqrt{n}/\|y\|$, the test (16) now reads

$$\max\left\{\left|\theta^\top x_j'\right| \,:\, \|\theta\| \leq 1, \quad \underline{\gamma} \leq \langle y', \theta\rangle \leq \overline{\gamma}\right\} < \frac{\sqrt{n}\lambda}{\|X_j\|}.$$

Lemma 4 concludes the proof. □

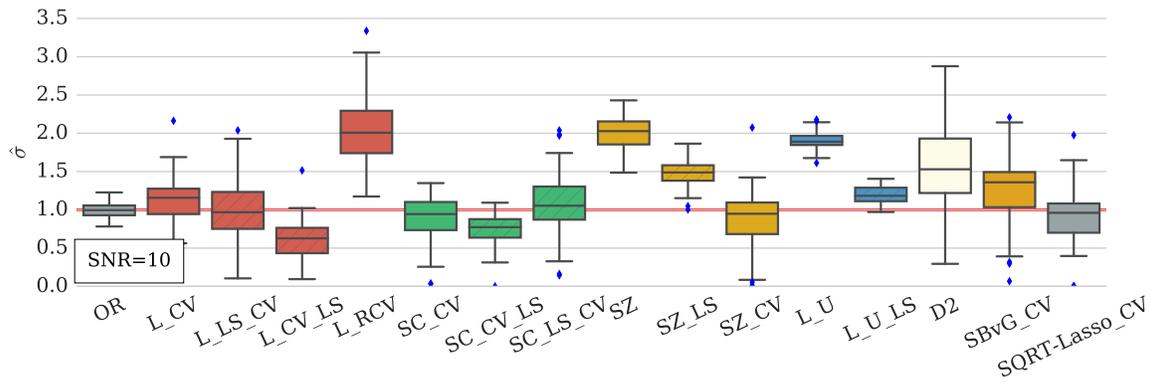

(a) $(n = 100, p = 200, \rho = 0, snr = 10, s = 0.8)$

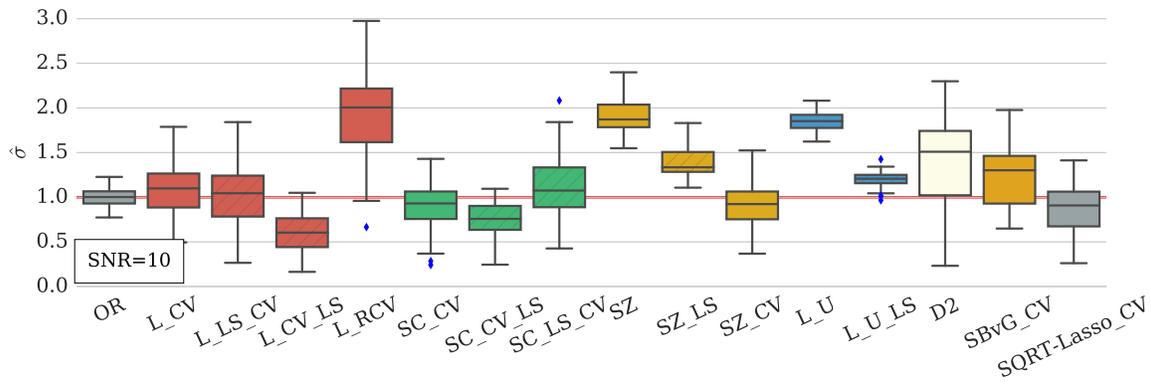

(b) $(n = 100, p = 200, \rho = 0.2, snr = 10, s = 0.8)$

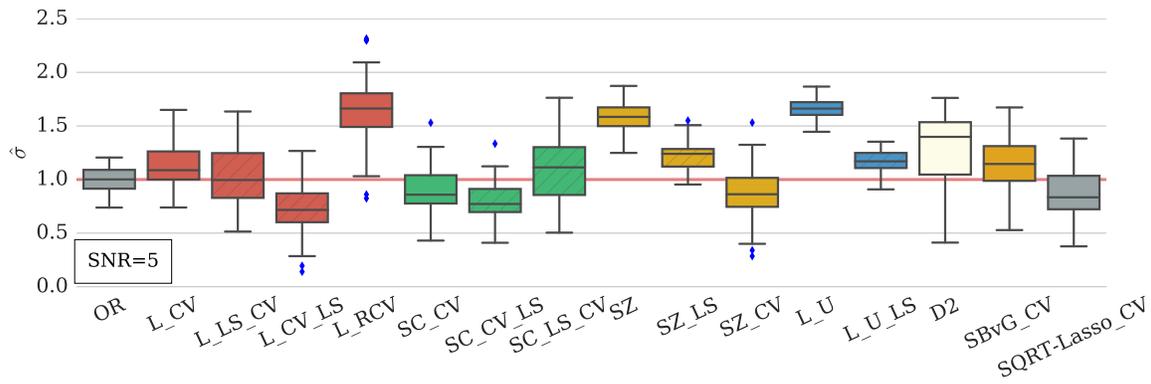

(c) $(n = 100, p = 200, \rho = 0.6, snr = 5, s = 0.8)$

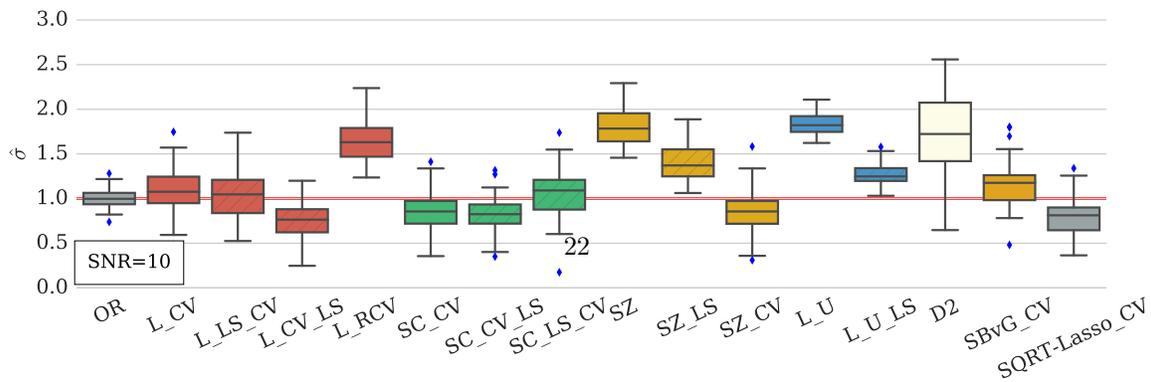

(d) $(n = 100, p = 200, \rho = 0.8, snr = 10, s = 0.8)$

Figure 3: Estimation performance on synthetic dataset.

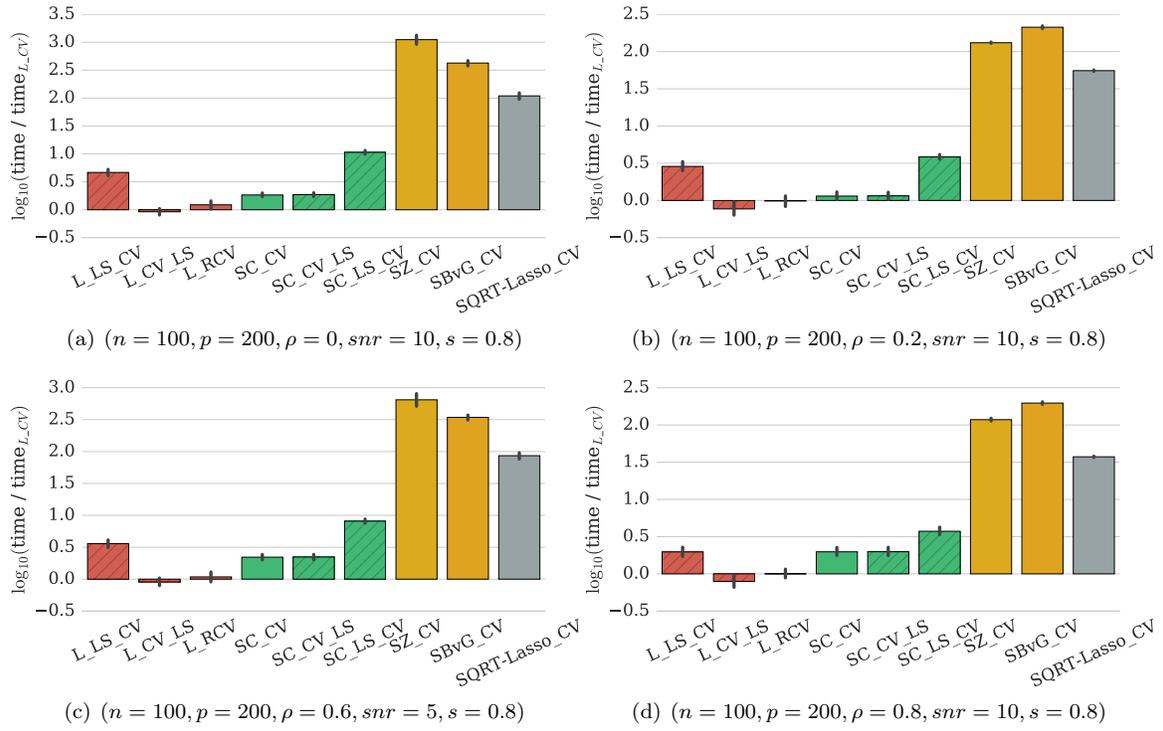

Figure 4: Computational time for 50 simulations on synthetic dataset.

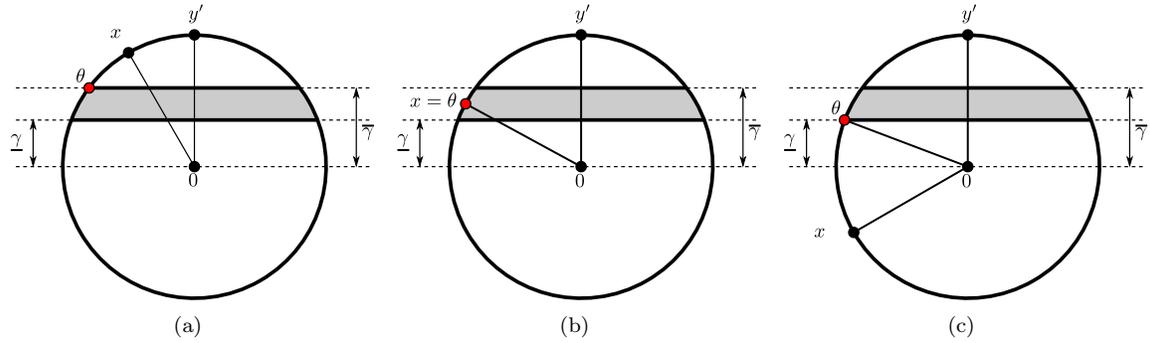

Figure 5: Three regimes for the optimal dual value. The grey region represents the intersection between the ball $\{\theta \in \mathbb{R}^n : \|\theta\| \leq 1\}$ and the set $\{\theta \in \mathbb{R}^n : \underline{\gamma} \leq \theta^\top y' \leq \overline{\gamma}\}$.

23